\documentclass[conference]{IEEEtran}
\IEEEoverridecommandlockouts
\usepackage{cite}
\usepackage{booktabs}
\usepackage{amsmath,amssymb,amsfonts}
\usepackage{algorithmic}
\usepackage{graphicx}
\usepackage{textcomp}
\usepackage{xcolor}
\usepackage{orcidlink}
\def\BibTeX{{\rm B\kern-.05em{\sc i\kern-.025em b}\kern-.08em
    T\kern-.1667em\lower.7ex\hbox{E}\kern-.125emX}}
\begin{document}

\title{SETU: An Agentic Ecosystem for Multilingual, Persona-Aware Communication Coaching
}

\author{
\IEEEauthorblockN{
Jonnalagadda Maruthi Tejas\textsuperscript{1}\orcidlink{0009-0007-9200-2167},
Uponika Barman Roy\textsuperscript{2},
Tilottama Goswami\textsuperscript{3}\orcidlink{0000-0003-0665-5190},
Samir Goswami\textsuperscript{1},
Mousita Dhar\textsuperscript{1}
}

\IEEEauthorblockA{
\textsuperscript{1}Quanta People Solutions Pvt. Ltd., Hyderabad, India
\quad
\textsuperscript{2}Panorama AI Solutions, London, Canada
}

\IEEEauthorblockA{
\textsuperscript{3}University College of Engineering, Osmania University, Hyderabad, India
}

\IEEEauthorblockA{
jmtejas.work@gmail.com,\quad uponika@panoramaai.org,\quad
goswami.tilottama@gmail.com
\\
samirgoswami@tminetwork.com,\quad mousitad@quantapeople.com
}
}

\maketitle

\begin{abstract}
Corporate training teams need scalable and explainable tools to improve workforce communication in multilingual settings. Existing systems often score text, audio, or video in isolation, or produce black-box outputs that are difficult to audit for coaching use. This paper presents SETU, an agentic ecosystem for \textit{corporate communication coaching} aimed at \textit{recruiters, frontline sales professionals and training units} who prepare for audience-specific conversations. SETU is designed for two scoped scenarios: (i) recruiter-candidate eligibility-and-interest calls with persona context and (ii) sales pitches with target-audience adaptation; owing to limited evaluation resources, this paper reports results on scenario (ii) only. The ecosystem decomposes analysis into specialized video, audio--speech, text--relevance, scoring, notification and reporting agents coordinated through trust-aware orchestration. It generates modality-attributed coaching reports for formative training, with human reviewers retaining final judgment. The name SETU (``bridge'' in several Indic languages) reflects the goal of bridging communication gaps across regional languages and audience expectations.
\end{abstract}

\begin{IEEEkeywords}
multimodal AI, agentic AI, communication coaching, multilingual speech, persona-based analysis
\end{IEEEkeywords}

\section{Introduction}
Professional communication in high-stakes workplace settings is a multimodal problem. Trainees are evaluated through what they say, how clearly they say it and how well they adapt to a specific audience. This challenge appears in recruitment calls, sales conversations and client-facing presentations, where the same message must be tuned to listener role, intent and context.

In multilingual Indian workplace settings, trainees may speak in English, Hindi, Telugu, or code-mixed forms such as English--Bengali. Existing coaching tools often provide generic textual feedback or single-modality scoring. They rarely combine visual, acoustic, linguistic and persona-aware reasoning in one auditable pipeline. Users therefore receive incomplete guidance on how to improve for a particular audience or task.

Recent progress in large language models (LLMs), speech processing and multimodal learning has improved communication support systems. PresentCoach \cite{presentcoach2025} uses dual-agent presentation coaching, while PersoPilot \cite{persopilot2026} shows the value of persona-aware personalization. However, most prior systems are monolithic. They do not decompose coaching into specialized, replaceable and explainable units. This limits debugging, modality attribution and governance in enterprise training workflows.

This paper presents SETU, an agentic ecosystem for multilingual interviews and persona-aware communication coaching. SETU is designed for corporate training and assessment platforms that require transparent evaluation of spoken communication. Beyond recruitment, the ecosystem can support sales profiling, needs gathering and product mapping. In this study, evaluation is scoped to the sales-pitch scenario; recruiter--candidate interactions are supported by design but are not evaluated here owing to limited resources.

An \textit{agentic ecosystem} is adopted rather than a single end-to-end model for four reasons. First, each modality has different failure modes and update cycles. Second, agent boundaries make it clear which component produced each score. Third, one agent can be replaced without retraining the full system. Fourth, trust-weighted orchestration allows unreliable signals (e.g., noisy audio) to be down-weighted during aggregation.

The work has the following four contributions:
\begin{enumerate}
    \item A multimodal agentic ecosystem that decomposes communication coaching into video, audio--speech, text--relevance, scoring, notification and reporting agents.
    \item A persona-aware analysis layer that adapts feedback to performer role, target audience, task prompt and multilingual language conditions.
    \item A trust-aware orchestration formulation for routing subtasks and aggregating modality evidence into structured coaching reports.
    \item A human-in-the-loop design that positions SETU as a formative coaching assistant rather than an autonomous hiring or sales decision system.
\end{enumerate}

The rest of this paper is organized as follows. Section~II reviews related work. Section~III describes the SETU agentic ecosystem and sales pitch mapping methodology. Section~IV presents the experimental case study and results. Sections~V and~VI discuss limitations and future work. Section~VII concludes the paper.
\section{Related Work}
AI-supported sales and presentation coaching draws from presentation simulation, multimodal assessment, multilingual speech processing, persona-aware interaction and responsible AI for high-stakes evaluation.

PresentCoach \cite{presentcoach2025} supports presentation practice through exemplars and interactive feedback. It is not designed for multilingual sales pitch mapping with buyer-persona relevance scoring. PersoPilot \cite{persopilot2026} improves persona-aware response generation, but it does not jointly analyze posture, prosody and code-mixed pitch delivery in a modular coaching workflow. Agentic multimodal advertising frameworks \cite{admultimodal2025} show persona-targeted reasoning, yet they target campaign outcomes rather than formative sales coaching. Low-resource language research \cite{lrla2025} highlights agentic opportunities for regional languages, but does not provide a deployable multimodal sales coaching stack. Work on code-switching automatic speech recognition (ASR) for Indic languages characterizes the recognition challenges that constrain multilingual coaching systems \cite{diwan2021asr,chadha2022codeswitched,liu2024aligning}.

LLM-as-judge surveys \cite{li2024llmjudge,gu2024judge} show that automated scoring is scalable but sensitive to prompt design, bias and calibration. Fairness studies in AI-driven assessment \cite{mujtaba2024fairness,mujtaba2025bias,biswas2024molly} warn that automated evaluation systems may amplify accent, language and appearance bias. These works motivate transparent rubrics and human oversight, but they do not by themselves provide a multimodal sales coaching pipeline.

Despite these advances, three gaps remain in the literature. First, many systems focus on generic presentation coaching without explicit sales-task and buyer-persona mapping. Second, persona-aware adaptation is often separated from multimodal behavioral analysis of spoken pitches. Third, multilingual and code-mixed sales speech is rarely treated as a first-class coaching dimension in workforce training systems.

The proposed SETU ecosystem addresses these limitations by integrating multimodal evidence, buyer-persona context, multilingual support, trust-aware orchestration and human oversight in one sales coaching pipeline.

SETU differs from these systems by coupling specialized agents, explainable scorecards and enterprise sales-training governance in a single agentic ecosystem.

\section{Methodology}
Building on the identified research gaps, this section presents the SETU agentic ecosystem for sales pitch mapping. It introduces the high-level architecture, mathematical formulation, pitch data design and agent workflow and uses a running payroll-SaaS pitch case study to show how the components connect in practice.

\subsection{Running Sales Pitch Case Study Setup}
One representative sales pitch session is used to illustrate the pipeline end-to-end:
\begin{itemize}
    \item \textbf{Scenario:} First-meeting product pitch to a prospective small and medium-sized enterprise (SME) buyer.
    \item \textbf{Performer:} Sales Performer pitching a cloud human resources (HR) and payroll software-as-a-service (SaaS) platform.
    \item \textbf{Target persona:} CEO of an SME company (focused on return on investment (ROI), time-constrained decision maker).
    \item \textbf{Pitch prompt:} ``In 90 seconds, convince the CEO to adopt your payroll automation platform. Cover the SME pain point, product value, ROI impact and a clear next step.''
    \item \textbf{Language:} English--Bengali code-mixed speech.
    \item \textbf{Input:} 78-second webcam pitch recording.
\end{itemize}
In the weak version of this pitch (Case A), the performer opens with company background, lists product modules and code-switches into Bengali without linking benefits to SME cost savings. In the strong version (Case B), the performer opens with payroll-compliance pain, cites a concrete ROI example and closes with a demo call-to-action. This pitch case is revisited in Subsections III-B to III-J.

\subsection{Proposed High-Level Agentic Framework}
\begin{figure*}[t]
\centering
\includegraphics[width=\textwidth]{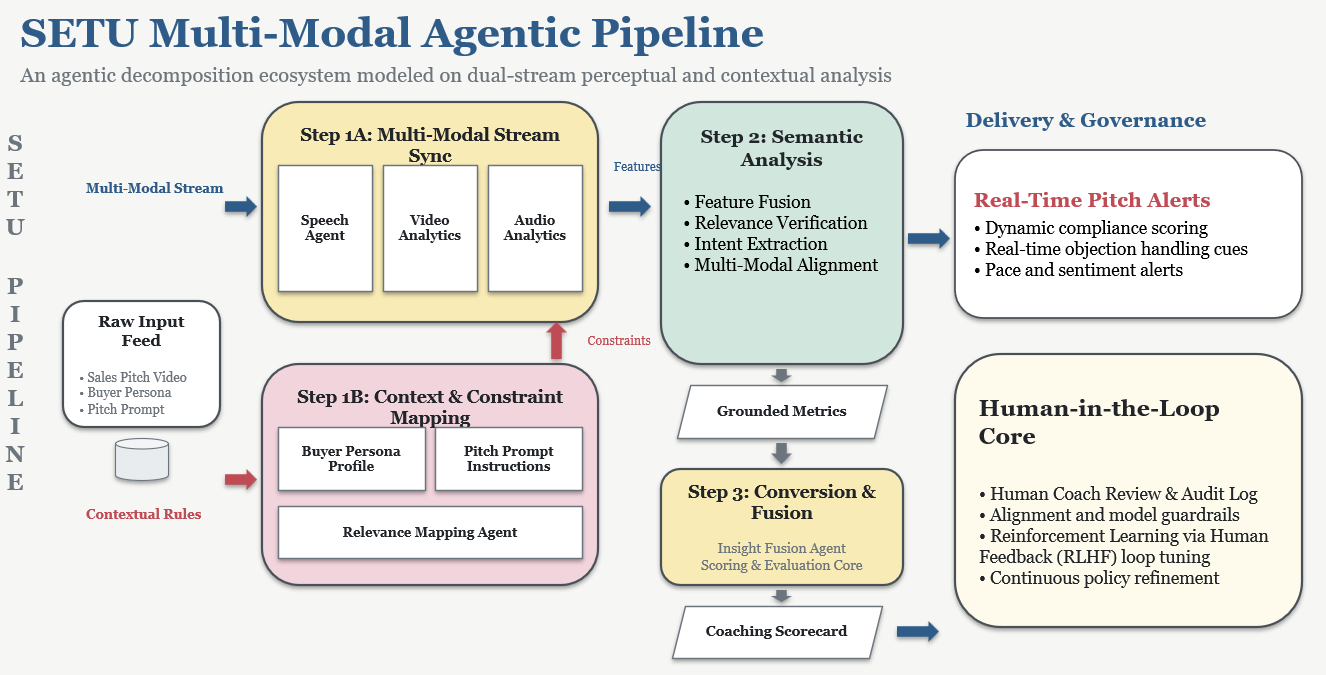}
\caption{SETU multimodal agentic pipeline for sales pitch mapping. The ecosystem uses dual-stream perceptual analysis (Step~1A) and contextual constraint mapping (Step~1B), followed by semantic fusion (Step~2), scoring and evaluation (Step~3) and delivery with human coach review.}
\label{fig:block}
\end{figure*}

Fig.~\ref{fig:block} shows the SETU multimodal agentic pipeline. The system uses a dual-stream design: Step~1A routes the synchronized pitch recording to the speech, video and audio agents, while Step~1B maps buyer persona and pitch-prompt constraints through the relevance mapping agent. Step~2 performs semantic analysis and multimodal alignment. Step~3 converts derived metrics into fused scores through the insight and evaluation core. The pipeline then delivers live pitch alerts and a coaching scorecard under human-in-the-loop governance.

In the pitch case study, the context agent injects CEO-of-SME expectations: concise delivery, business impact and measurable ROI. The speech agent transcribes the spoken pitch, including code-mixed segments. The relevance agent checks whether the pitch follows a persuasive structure (pain point $\rightarrow$ solution $\rightarrow$ ROI $\rightarrow$ close), not merely product feature listing.

\subsection{Mathematical Representation}
The multimodal agentic ecosystem is formulated as:
\begin{equation}
\mathcal{M} = (A,T,S,E,O,C,\mathit{Mem}).
\label{eq:system}
\end{equation}
Equation~(\ref{eq:system}) defines the tuple form of SETU. $A$ is the agent set, $T$ is the global coaching task, $S$ is state, $E$ is input environment, $O$ is orchestrator, $C$ is communication among agents and $\mathit{Mem}$ is memory.

The core modality agents are instantiated as:
\begin{equation}
A_{\mathrm{core}} = \{v,\,a,\,t\},
\label{eq:agents}
\end{equation}
where $v$ is the video agent, $a$ is the audio/speech agent and $t$ is the text/relevance agent group. Extended operational agents (sync, context, insight, scoring, notification, report) are included in the full set $A$.

The global task is:
\begin{equation}
T = \{t_1,t_2,t_3,t_4,t_5,t_6\}.
\label{eq:task}
\end{equation}
In (\ref{eq:task}), $t_1$ synchronizes streams, $t_2$ extracts visual cues, $t_3$ performs speech and acoustic analysis, $t_4$ evaluates text quality and language profile, $t_5$ computes persona-aware sales relevance and $t_6$ generates the coaching report.

The orchestration assignment is
\begin{equation}
O(T,S,A,\mathit{Mem}) \rightarrow \{(x_i,t_j)\},
\label{eq:orch}
\end{equation}
where $x_i \in A$ is the selected agent for subtask $t_j$. Equation~(\ref{eq:orch}) is the routing core of the ecosystem: each subtask is mapped to the most capable agent instead of sending all inputs to one monolithic model.

For example,
\begin{equation}
\begin{split}
O(T)=\{(sync,t_1),(v,t_2),(a,t_3),\\
(t,t_4),(t,t_5),(report,t_6)\}.
\end{split}
\label{eq:orch_example}
\end{equation}
Equation~(\ref{eq:orch_example}) means stream sync handles alignment, $v$ handles visual behavior, $a$ handles speech and acoustics, $t$ handles linguistic and relevance reasoning and the report agent compiles outputs. In the pitch case study, relevance is assigned to $t$ because it must judge whether the spoken pitch addresses the CEO prompt using both transcript evidence and buyer-persona context.

Each agent is represented as
\begin{equation}
x_i=(\pi_i,\phi_i,\psi_i),
\label{eq:agent_form}
\end{equation}
where $\phi_i$ is feature extraction, $\pi_i$ is decision policy and $\psi_i$ is structured output generation. Equation~(\ref{eq:agent_form}) separates perception, reasoning and structured output generation for every agent.

State evolution is
\begin{align}
s_0 &= \text{Raw Input}, \label{eq:s0}\\
s_1 &= f(s_0,sync), \label{eq:s1}\\
s_2 &= f(s_1,\{v,a\}), \label{eq:s2}\\
s_3 &= f(s_2,\{a,t\}), \label{eq:s3}\\
s_4 &= f(s_3,report). \label{eq:s4}
\end{align}
Equations~(\ref{eq:s0})--(\ref{eq:s4}) show progressive enrichment of context. For the pitch case study, $s_2$ adds posture and vocal-delivery features from the recorded pitch; $s_3$ adds transcript text and pitch-relevance evidence; $s_4$ is the final pitch coaching report.

Trust of agent $x_i$ is
\begin{equation}
\mathrm{Trust}(x_i)=\alpha\,\mathrm{Acc}_i+\beta\,\mathrm{Cons}_i+\gamma\,R_i,
\label{eq:trust}
\end{equation}
where $\mathrm{Acc}_i$ is modality accuracy, $\mathrm{Cons}_i$ is score consistency across repeated windows and $R_i$ is runtime reliability (latency failures, API dropout). Equation~(\ref{eq:trust}) is applied to \textit{agent outputs}, not to the sales trainee. In implementation, trust weights modality contributions during aggregation, e.g., down-weighting audio evidence when the signal-to-noise ratio (SNR) is poor.

The orchestration design criterion is
\begin{equation}
\max \sum_i R_i - \lambda C_{\mathrm{coord}},
\label{eq:objective}
\end{equation}
where $C_{\mathrm{coord}}$ is coordination cost (extra agent calls and delay). Equation~(\ref{eq:objective}) expresses the trade-off between output reliability and coordination cost.

\subsection{Sales Pitch Mapping Design and Data Acquisition}
The sales pitch mapping design is a structured training matrix for spoken product-pitch practice. Each entry defines \textit{what product to pitch}, \textit{to which buyer persona} and \textit{which pitch objective} must be achieved within the time limit.

The design contains four layers:
\begin{enumerate}
    \item \textbf{Performer role:} Sales Performer delivering the pitch.
    \item \textbf{Target buyer persona:} CEO of SME, dual-income, no-kids (DINK) professional in an information technology (IT) firm, Retired Government Officer, School Teacher, or Mid-level Manager.
    \item \textbf{Pitch stage:} Lead Generation, Connecting, Profiling, Needs Gathering, or Product Mapping.
    \item \textbf{Pitch prompt:} A timed instruction such as ``Pitch the payroll SaaS platform to an SME CEO in 90 seconds with ROI and next-step close.''
\end{enumerate}

Each matrix cell also specifies language condition and exemplar quality. Language conditions are English, Hindi, Telugu, English--Hindi, English--Bengali and English--Telugu. Exemplar bands are Good, Fair and Poor. Good/Fair/Poor pitch clips are used as mirroring references for personalized feedback, not as automatic pass/fail labels.

For pitch data collection, video quality is controlled at 720p webcam capture with single-speaker framing. Audio is recorded in typical office noise conditions. Pitch duration is 45--120 seconds per clip. Sales coaches mark each pitch as Good, Fair, or Poor based on pitch-specific rubrics: opening hook, buyer relevance, evidence/ROI, delivery quality and closing call-to-action.

In the running pitch example, the performer must convince an SME CEO that payroll automation reduces compliance risk and operating cost. The pitch should be persuasive and buyer-specific, not a generic product brochure delivered aloud.

\subsection{Video Agent}
The video agent ($v$) extracts posture stability, eye-contact approximation, facial expressiveness and head movement using MediaPipe Holistic landmarks. Sub-scores are temporally smoothed and combined into a visual performance score.

In Case A, the performer looked away from the camera while stating pricing and lost upright posture during the ROI segment. This produced a visual score of 3.8/10. The report noted that weak non-verbal delivery reduced perceived pitch confidence, independent of language choice.

\subsection{Speech and Audio Agents}
The audio/speech agent ($a$) performs two functions. First, it transcribes multilingual and code-mixed speech using the Sarvam speech-to-text (STT) model (\texttt{saaras:v3}, \texttt{codemix}). Second, it computes acoustic features with Librosa: root-mean-square (RMS) energy, pitch variation, silence ratio, SNR, words-per-minute (WPM) and filler density.

Language detection combines API confidence, script-based fallback and parallel language probes. This is important for Indian code-mixed sales speech where script cues and audio cues may disagree.

In Case A, the pitch transcript contained frequent ``um/uh'' pauses before key value statements and Bengali code-switching appeared during the ROI explanation. English and Bengali were detected. Filler density was high and SNR was moderate, yielding audio score 4.1/10 with a coaching alert to slow down and emphasize benefit statements clearly.

\subsection{Text Agent}
The text agent ($t$) evaluates grammar, fluency, professionalism and semantic completeness using rubric-constrained LLM scoring. It compares the transcript against the sales-task prompt and buyer persona context. A performer is scored for delivery quality and for addressing the correct sales communication objective.

Code-switching is not penalized by default. The agent flags excessive language switching only when more than three languages appear in one session (English plus up to two regional languages is the coaching target).

In Case A, the performer said: ``Our company was founded in 2018... we have attendance, payroll, leave modules...'' without converting features into buyer outcomes. Text score was 4.6/10, with feedback on fragmented pitch flow and missing value proposition language.

\subsection{Relevance Agent}
The relevance agent evaluates whether the pitch satisfies the persona-task prompt. It enriches the prompt with performer and target buyer descriptions before LLM rubric scoring on a 0--50 scale, normalized to 0--10.

In Case A, the pitch did not answer the CEO prompt. It omitted payroll-compliance pain, offered no ROI figure and ended without a next-step ask. Relevance was partial (5.4/10), with deductions for missing pain point, missing ROI evidence and missing close. By contrast, Case B opened with ``Manual payroll errors cost SMEs time and penalties,'' cited ``up to 30\% admin-time reduction,'' and closed with ``Can we schedule a 15-minute demo Thursday?''

\subsection{Orchestrator}
The orchestrator realizes the routing and trust-weighted aggregation of Equations~(\ref{eq:orch})--(\ref{eq:trust}) in a LangGraph workflow with shared \texttt{AgentState}. It routes subtasks, stores intermediate outputs and applies trust-weighted aggregation when modality reliability is unequal.

For the case study, audio trust was reduced due to noise, while text/relevance trust remained stable. This prevented noisy acoustics from dominating the final score.

\subsection{Overall Coaching Report and Recommendations}
The session report includes visual, audio, text, relevance, language profile, overall score and recommendations. Each recommendation is linked to modality evidence and deduction traces.

For Case A, the pitch coaching report recommended a four-part rewrite: (i) open with SME payroll pain, (ii) present two buyer-relevant benefits with one ROI metric, (iii) maintain eye contact during price/value statements and (iv) close with a concrete demo call-to-action. It also advised keeping English--Bengali switching only where it improves rapport, not during core value delivery.

\subsection{Human in the Loop}
SETU is a coaching assistant, not an autonomous decision engine. Human sales coaches validate reports, inspect modality evidence, adjust rubrics and approve final feedback before trainee delivery.

In enterprise use, this human validation step supports auditability and fairness governance.

\section{Experiments and Performance Analysis}
After describing the SETU agentic ecosystem, this section evaluates it through a sales pitch mapping case study. It reports the pitch evaluation setup, quantitative results, comparative baselines and coaching-time impact on sales trainers.

\subsection{Case Study Scope and Objectives}
SETU is evaluated through a pilot case study with a corporate sales training partner. The evaluation is scoped to \textit{sales pitch mapping only}. Every session is a timed product pitch: a sales performer presents a solution to a defined buyer persona within 45--120 seconds.

Pitch stages include Lead Generation, Connecting, Profiling, Needs Gathering and Product Mapping. However, each recording is evaluated as a persuasive pitch, not as a free-form conversation. The core question is: ``Did the spoken pitch fit the buyer persona and achieve the prompt objective?''

Objectives are to measure (i) pitch delivery scoring validity, (ii) multilingual pitch handling quality, (iii) buyer-persona pitch relevance, (iv) explainability of pitch feedback and (v) coaching-time reduction for human sales trainers.

\subsection{Dataset Characteristics}
The pilot dataset contains 18 recorded sales pitch videos. Each video captures one complete pitch attempt against a fixed prompt:
\begin{itemize}
    \item \textbf{Products pitched:} HR--payroll SaaS (8), EdTech learning platform (5), insurance cross-sell plan (5).
    \item \textbf{Target personas:} CEO of SME (6), School Teacher (4), Mid-level Manager (4), DINK in IT firm (2), Retired Government Officer (2).
    \item \textbf{Pitch stages:} Lead Generation (4), Connecting (3), Profiling (3), Needs Gathering (4), Product Mapping (4).
\end{itemize}
Pitch length is 45--120 seconds (mean 78 seconds, SD 19 seconds). Resolution is 720p webcam capture. Audio is mono microphone input in office environments. Language distribution: English (4), Hindi (3), Telugu (2), English--Hindi (5), English--Telugu (4). Coaches label each pitch as Good, Fair, or Poor using pitch rubrics (hook, relevance, evidence, delivery, close).

\subsection{Evaluation Protocol}
Each video is processed by the SETU batch pipeline: audio extraction, STT, language profiling, video analytics, acoustic analytics, LLM rubric scoring and weighted aggregation. Live-mode tests are run on a 6-session subset to measure alert latency.

Baselines:
\begin{itemize}
    \item \textbf{B1:} Text-only LLM feedback (transcript-only).
    \item \textbf{B2:} Non-agentic multimodal prompt (single-pass summary without agent-level traces).
\end{itemize}

\subsection{Metrics}
The evaluation reports modality sub-scores (0--10), overall score, coach agreement (Spearman $\rho$), explainability rating (coach 1--5), language macro-F1, end-to-end latency and coaching-time reduction.

\subsection{Representative Pitch Case Study Results}
Table~\ref{tab:case_results} shows three representative pitch sessions, including the running methodology example (Case A).

\begin{table}[t]
\centering
\caption{Representative sales pitch mapping scores (0--10)}
\label{tab:case_results}
\begin{tabular}{p{0.34\linewidth}cccc}
\toprule
\textbf{Pitch Session} & \textbf{Vis.} & \textbf{Aud.} & \textbf{Text} & \textbf{Rel.} \\
\midrule
Case A: Weak payroll SaaS pitch to SME CEO (English--Bengali) & 3.8 & 4.1 & 4.6 & 5.4 \\
Case B: Strong payroll SaaS pitch to SME CEO (English--Bengali) & 7.6 & 7.3 & 7.8 & 8.1 \\
Case C: EdTech pitch to School Teacher (English) & 6.9 & 6.5 & 7.2 & 8.4 \\
\bottomrule
\end{tabular}
\end{table}

Overall scores were 4.6 (Case A), 7.7 (Case B) and 7.4 (Case C) using batch weights $0.20$ visual, $0.25$ audio, $0.20$ text, $0.35$ relevance.

Across all 18 sales sessions, SETU obtained a mean batch latency of 38.6 s per clip, live alert refresh 2--4 s, coach agreement $\rho=0.78$, explainability 4.3/5 and language macro-F1 0.86.

\subsection{Comparative Analysis}
SETU outperformed B1 ($\rho=0.61$, explainability 2.9/5) and B2 ($\rho=0.69$, explainability 3.4/5) on the same pilot set. Buyer-persona-conditioned relevance improved agreement by 0.11 in $\rho$ over non-persona prompting.

Coach review time was reduced from 22 to 13 minutes per session (40.9\% reduction) and the average number of practice cycles needed to reach Fair/Good decreased from 4.1 to 2.6.

\subsection{Case Study Insights}
The sales pitch case study shows that multimodal agent decomposition produces pitch feedback that coaches can act on immediately. Buyer-persona context is especially important for relevance scoring: the same product language can score high for a School Teacher pitch but low for an SME CEO pitch if ROI is missing. Code-mixed pitch delivery is feasible to coach when ASR, script cues and confidence filtering are combined. Human coach review remains necessary to validate pitch judgments and avoid over-penalizing natural language switching.

\section{Limitations}
Although the pilot case study provides initial evidence, SETU has important boundaries that must be stated before broader deployment. This section summarizes the main technical, data and governance constraints observed in the current study.

SETU depends on video/audio quality, ASR accuracy, prompt design and rubric fairness. Visual cues such as eye contact may vary with culture, disability, camera angle and lighting. The current pilot is small (18 sales videos) and internal. Results should be validated on larger expert-labeled benchmarks before deployment at scale.

\section{Future Work}
Based on the case study findings and the limitations above, this section outlines the next steps needed to strengthen SETU for large-scale sales training use.

Future work will expand the sales dataset across more buyer personas and task categories, add expert-labeled benchmarks, improve low-resource language support, add longitudinal coaching analytics and conduct fairness audits across gender, accent, skin tone, disability and regional language groups.

\section{Conclusion}
This paper examined whether an agentic ecosystem can provide explainable, persona-aware coaching for multilingual sales pitches.

This paper presented SETU, an agentic ecosystem for multilingual sales pitch mapping and persona-aware communication coaching. By combining specialized agents, buyer-persona context, trust-aware orchestration and human oversight, SETU provides explainable and specific coaching for sales communication training. Case-study evaluation on multilingual sales pitch scenarios indicates improved coach agreement and reduced pitch-coaching effort compared with text-only and non-agentic baselines.

\section*{Acknowledgment}
The authors thank Quanta People Solutions Pvt. Ltd. for supporting this research work.

\end{document}